\documentclass[sigconf]{acmart}

\usepackage{hyperref}
\usepackage{tabularx}
\usepackage{caption}
\usepackage{diagbox}
\usepackage{longtable}
\usepackage{multirow}
\usepackage{booktabs}
\usepackage{subcaption}
\usepackage{graphicx}
\usepackage{xcolor}
\usepackage[figuresright]{rotating}

\AtBeginDocument{%
  }

\setcopyright{acmlicensed}
\copyrightyear{2024}
\acmYear{2024}
\acmDOI{10.1145/3696409.3700165}

\acmConference[MMASIA '24]{ACM Multimedia Asia}{December 3--6, 2024}{Auckland, New Zealand}

\acmBooktitle{ACM Multimedia Asia (MMASIA '24), December 3--6, 2024,
Auckland, New Zealand}
\acmISBN{979-8-4007-1273-9/24/12}

\begin{document}

\title{Fire and Smoke Detection with Burning Intensity Representation}


\author{Xiaoyi Han}
\orcid{0009-0000-2300-4264}
\affiliation{%
  \institution{School of Software Technology, Zhejiang University}
  \city{}
  \country{}
}
\author{Yanfei Wu}
\orcid{0009-0006-7551-565X}
\affiliation{%
  \institution{China Mobile (Suzhou) Software Technology Co., Ltd.}
  \city{}
  \country{}
}
\author{Nan Pu}
\orcid{0000-0002-2179-8301}
\affiliation{%
  \institution{University of Trento}
  \city{}
  \country{}
}
\author{Zunlei Feng}
\orcid{0000-0001-8640-8434}
\affiliation{%
  \institution{School of Software Technology, Zhejiang University}
  \city{}
  \country{}
}
\author{Qifei Zhang}
\orcid{0009-0001-8247-4562}
\affiliation{%
  \institution{School of Software Technology, Zhejiang University}
  \city{}
  \country{}
}
\author{Yijun Bei}
\orcid{0000-0001-5720-6374}
\affiliation{%
  \institution{School of Software Technology, Zhejiang University}
  \city{}
  \country{}
}
\author{Lechao Cheng}
\orcid{0000-0002-7546-9052}
\authornotemark[1]
\email{chenglc@hfut.edu.cn}
\affiliation{%
  \institution{Hefei University of Technology}
  \city{}
  \country{}
}

\renewcommand{\shortauthors}{Xiaoyi Han. et al.}
\begin{abstract}
An effective Fire and Smoke Detection (FSD) and analysis system is of paramount importance due to the destructive potential of fire disasters. However, many existing FSD methods directly employ generic object detection techniques without considering the transparency of fire and smoke, which leads to imprecise localization and reduces detection performance. To address this issue, a new Attentive Fire and Smoke Detection Model (a-FSDM) is proposed. This model not only retains the robust feature extraction and fusion capabilities of conventional detection algorithms but also redesigns the detection head specifically for transparent targets in FSD, termed the Attentive Transparency Detection Head (ATDH). In addition, Burning Intensity (BI) is introduced as a pivotal feature for fire-related downstream risk assessments in traditional FSD methodologies. Extensive experiments on multiple FSD datasets showcase the effectiveness and versatility of the proposed FSD model. The project is available at \href{https://xiaoyihan6.github.io/FSD/}{https://xiaoyihan6.github.io/FSD/}.\footnote{Code is available: \href{https://github.com/XiaoyiHan6/FSDmethod}{https://github.com/XiaoyiHan6/FSDmethod}.}
\end{abstract}






\maketitle

\section{Introduction}
Fire, a formidable and highly perilous disaster, can result in substantial loss of life, property, societal upheaval, and adverse ecological consequences across diverse scenes, such as urban conflagrations~\cite{1}, forest fires~\cite{2}, industrial infernos~\cite{3}, and transportation incidents~\cite{4}. Considering the profound destructive potential of fire disasters, investigating and developing automated Fire and Smoke Detection (FSD) and analysis systems is of significant importance and is urgently needed~\cite{5,pu2023dynamic,pu2023memorizing,49}. Furthermore, approaches that employ generic object detection methods in FSD without taking into account the transparency of fire and smoke undermine the overall performance of FSD methods and may even result in instances of missed detection and false alarms. Additionally, generic object detection algorithms currently lack an effective linking mechanism for seamless implementation in downstream wildfire risk assessment tasks.\\
\indent
It is found that transparent flames and smoke frequently appear during fire disasters due to the distinctive features of combustible materials, the environmental conditions of the fire, and the unique optical properties of certain substances. This transparency effect can cause indistinct boundaries between the flame or smoke targets and the background in FSD scenarios, ultimately degrading the performance of FSD. In addition, the phenomenon may further lead to missed or false detection of flames or smoke, posing a challenge to the accuracy of FSD. To mitigate these drawbacks, unlike existing methods~\cite{6} that directly apply generic object detection algorithms to FSD, a new Attentive Transparency Detection Head (ATDH) is proposed, which is designed for adaptive enhancement of transparency features across multiple scales and channels. ATDH is compatible with most generic object detection algorithms to enhance the accuracy of FSD.\\
\indent
Moreover, a Relative Burning Intensity (BI) is defined to represent BI. To the best of our knowledge, this is the first representation of BI. This design serves as an indicator to measure the severity of combustion and endows generic FSD algorithms with a novel and essential ability for fire-related risk assessment, which plays a crucial role in determining the allocation of firefighting resources, developing effective fire management strategies, and improving the efficiency of fire suppression efforts.\\
\indent
In summary, the main~\textbf{contributions} are as follows: 1) A novel \textbf{Attentive Fire and Smoke Detection Model (a-FSDM)} is proposed. Retaining the advantages of generic detection algorithms in feature extraction and fusion, the fire-related target detection head network is redesigned to meet the distinctive needs of FSD. 2) A new \textbf{Attentive Transparency Detection Head (ATDH)} is presented, which is tailor-made to address the challenge of detecting transparent fire and smoke targets by enhancing the distinct features of transparent targets while suppressing non-target feature maps. 3) \textbf{Burning Intensity (BI)} is introduced as an indicator to measure the severity of combustion, and it serves as a key feature for subsequent downstream assessment of fire damage.\\
\indent

\section{Related Work}
Fire and Smoke Detection (FSD) technology can be categorized into two types: smoke detection and flame (fire) detection. Smoke detection, which utilizes the characteristics of smoke, typically allows for earlier detection, thereby minimizing resource and economic losses. However, it may face challenges in environments where smoke disperses slowly or is not visible. Conversely, flame detection relies on the distinct features of flames, offering more easily detectable signals but often occurring at a later stage, when the fire is already more developed~\cite{7}.\\
\indent
In terms of FSD task classification, the model described in~\cite{8} involves training pre-trained VGG16~\cite{9} and ResNet50~\cite{10} models using a custom FSD dataset~\cite{8} to enhance FSD performance. Afterwards, a novel lightweight classification network designed for fire incidents is proposed by FireNet~\cite{11}. Subsequently, FireNet-v2, an improved version of FireNet, achieves a percision of 94.95~\cite{12}. Moreover, a recent study~\cite{13} introduces a deep learning network that combines CNN and RNN~\cite{14} for the purpose of forest fire classification. Subsequently, a novel smoke recognition method based on dark-channel assisted mixed attention~\cite{15}, is proposed. However, the traditional classified FSD task is limited to determining the existence of fire, which cannot provide more valuable information, such as fire location, for firefighters.\\
\indent
In recent years, various detection algorithms are incorporated with the aim of enhancing FSD task systems. For example, Faster RCNN~\cite{16} is introduced in~\cite{17}. To reduce false and missed detections, additional algorithms such as SSD~\cite{18}, R-FCN~\cite{19}, and YOLOv3~\cite{20} are integrated by researchers~\cite{21}, resulting in high levels of accuracy in their respective datasets. Li et al.~\cite{23} demonstrate the successful application of the widely-used DERT~\cite{22} to FSD tasks. Meanwhile, the FSD network GLCT~\cite{24}, which merges CNN and Transformers~\cite{25}, achieves an mAP of 80.71 in the detection of early fire in surveillance video images. However, despite their effectiveness, Transformers have higher computational resource requirements and slower inference times than traditional CNNs. Subsequently, Venâncio et al.~\cite{27} introduce YOLOv5~\cite{26} to FSD tasks, reporting improved accuracy with a smoke detection accuracy of 85.88 and a flame detection accuracy of 72.32 on their surveillance video database. It is noteworthy that traditional object detection-based FSD models are constructed upon general object detection algorithms, rather than being specifically designed for the FSD task.\\
\indent
Deep learning techniques are applied to image segmentation~\cite{28} and scene understanding~\cite{29} in FSD tasks. Guan et al.~\cite{30} develop MaskSU RCNN, a forest fire instance segmentation approach based on the MS RCNN model. Meanwhile, Perrolas et al.~\cite{31} propose a quad-tree search-based method for localizing and segmenting fires at different scales. It is noteworthy that fire and smoke in early-stage FSD present considerable difficulties due to their transparency and lack of specificity. The complex and variable nature of fire and smoke results in suboptimal performance of traditional semantic segmentation techniques~\cite{32}. Moreover, in comparison to semantic segmentation models that require substantial computational resources for training and inference, detection algorithms are more appropriate for FSD tasks.\\
\indent
Despite advancements in FSD tasks using object detection methods, computer vision-based FSD algorithms still lag behind generic computer vision algorithms. For example, the detection of transparent foregrounds has rarely been addressed in previous FSD studies. Moreover, generic object detection does not adequately address subsequent fire-related concerns, including BI, which is crucial for evaluating the cost of fire damage and determining the human resources required to respond to a fire. \\
\indent
In order to address the issue of the transparent foreground, a novel FSD method, namely the Attentive Fire and Smoke Detection Model (a-FSDM), is presented. The proposed method preserves the strengths of traditional detection algorithms in feature extraction and fusion while redesigning the target detection head network specifically for FSD, termed the Attentive Transparency Detection Head (ATDH). Furthermore, it is assumed that there is a positive correlation between the severity of fire and smoke disasters and Burning intensity (BI). This indicates that higher BI levels correspond to greater impacts caused by these disasters. To this end, a novel representation of BI is developed with the aim of facilitating evaluation for subsequent downstream tasks related to FSD. Additionally, the proposed algorithm is compared with multiple multi-scale object detection baselines on various FSD datasets.\\
\indent

\begin{figure*}
  \includegraphics[width=1\textwidth]{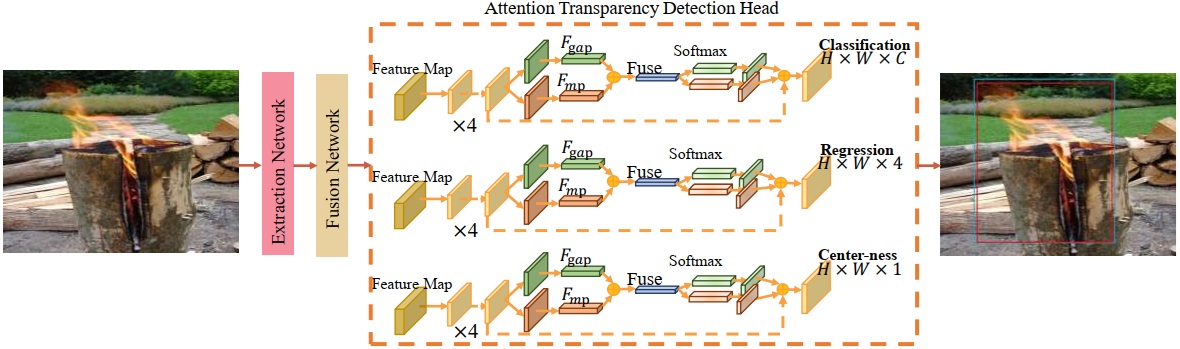}
  \vspace{-1em}
  \caption{The framework of a-FSDM. It retains three parts: the feature extraction and fusion networks of the general object detection algorithm, and a novel ATDH.}
  \vspace{-1em}
  \Description{}
  \label{fig1}
\end{figure*}

\section{Method}
The proposed method contains two parts: the Attentive Fire Smoke Detection  Model (a-FSDM) and the Representation of Burning Intensity (BI). In Fig.~\ref{fig1}, the a-FSDM is presented, and it consists of three main components: Feature Extraction, Feature Fusion Grouping, and the Attentive Transparency Detection Head (ATDH) for the FSD task. Feature Fusion Grouping is used to fuse semantic and spatial information from various convolutional layers to prevent information loss. The ATDH is responsible for classification, regression, and centerness.\\
\indent

\subsection{Attention Transparency Detection Head}
To enhance the model's comprehension and detection of transparent flame or smoke targets, a new Attentive Transparency Detection Head (ATDH) is proposed. The feature map output from the Neck undergoes four convolutional layers, followed by Global Average Pooling (GAP) and Max Pooling (MP). GAP ensures that higher values contribute more to the overall training, while MP assigns importance to the maximum value~\cite{33}. To avoid downplaying the significance of the maximum value, MP is introduced. The resulting feature maps from the two pooling methods are then fused, summarizing information from multi-scale features. Subsequently, softmax normalization is applied to generate a normalized score for each channel, guiding the model in assigning representation to different focus kernels~\cite{34}. To avoid network degradation and the vanishing gradient problem, a shortcut connection~\cite{10} is established between the feature map of this layer and the feature map of the attention mechanism. This connection compensates for the multiple convolution operations and layers that the feature map undergoes. Finally, the detection head provides confidence, location information, and centerness score.\\
\indent

\subsection{Representation of Burning Intensity}
In practice applications, Burning Intensity (BI) is a complex outcome influenced by various physical and chemical processes, including factors such as flame height, temperature, thermal radiation, fuel consumption rate, fuel product formation rate, and the composition and concentration of combustion products. Despite this complexity, to minimize costs and ensure errors remain within an acceptable range, the Intersection over Union (IoU) and Area Difference Normalization are considered as Relative RBI in its ideal state.\\
\indent
\noindent Specifically, considering the predicted coordinates \footnote{$d=(d^{cx},d^{cy},d^{w},d^{h})$}, and the ground truth coordinates \footnote{$g=(g^{cx},g^{cy},g^{w},g^{h})$}. Additionally, the area of the prediction box can be determined $A_d=d^w \times d^h$, and the area of the ground truth box $A_{g}=g^{w} \times g^{h}$. Therefore, Area Difference Normalization ${\mathop{\tilde{A}}}^{\prime}$ can be expressed as shown in Eq.~\ref{eq1}:\\
\indent
\vspace{-1em}
\begin{equation}
   {\mathop{\tilde{A}}}^{\prime}  =1-\left |\frac{A_{d}-A_{g}}{max(A_{d},A_{g})} \right|.
   \label{eq1}
\end{equation}

\noindent The value of ${\mathop{\tilde{A}}}^{\prime}$ approaches 1 when the prediction box and the ground truth area have the same area and approaches 0 when they do not overlap.\\
\indent
Consequently, the Relative BI of Fire or Smoke is determined using the IoU of the prediction box area and ground truth area, combined with ${\mathop{\tilde{A}}}^{\prime}$, represented as $\tilde{BI}$ in Eq.~\ref{eq2}:\\
\indent
\vspace{-1em}
\begin{equation}
  \tilde{BI} = w_{1} \times {\mathop{\tilde{A}}}^{\prime}  + w_{2} \times IoU, 
  \label{eq2}
\end{equation}

\noindent where $w_{1}$ and $w_{2}$ can be allocated equal weights of 1/2 to IoU and ${\mathop{\tilde{A}}}^{\prime}$ respectively, to ensure a balanced total weight of 1. If an FSD model consistently underperforms on small targets, leading to a significant discrepancy between the prediction box and the ground truth box, one strategy to address this issue is to amplify the importance of $ {\mathop{\tilde{A}}}^{\prime} $. Assigning a higher weight to $w_{1}$, effectively penalizes the model in case of inaccurate predictions on smaller fire and smoke targets. Conversely, when the model exhibits high accuracy in target localization, as evidenced by a high IoU score, but still displays minor inaccuracies in area prediction, prioritizing $w_{2}$ by assigning it a higher weight may be a suitable choice. This approach highlights and rewards the model's strong performance in target localization, even in scenes where slight biases in area predictions exist.\\
\indent
It is important to note that BI is expressed as a relative value, not an absolute one. To apply this data in real-world scene simulations, precise knowledge of camera sensor parameters is essential. However, variations in focal length exist in FSD datasets sourced from online collections. Neglecting these differences and using a single focal length for real-world projections can introduce significant errors. Nonetheless, the current relative error falls within an acceptable range for the argumentative requirements of FSD tasks. The BI indicator reflects the severity of fires, providing support for downstream assessment tasks related to fire incidents.\\
\indent

\section{Experiments}
\subsection{Setting and Details}
\textbf{Datasets.} Experiments are conducted on multiple public FSD datasets.\\
\indent
\textbf{Fire-Smoke-Dataset}~\cite{39} comprises 3,000 images. It contains 2,100 training images and 900 testing images, and the location of this dataset is annotated.\\
\indent
\textbf{Furg-Fire-Dataset}~\cite{40} consists of 23 videos. These videos are divided into individual frames to form images. The computer randomly deletes redundant photos while retaining 3,000 images, which are then labeled accordingly.\\
\indent
\textbf{VisiFire}~\cite{41} is a public FSD dataset consisting of 40 videos. The processing principle remains consistent with the earlier method mentioned, resulting in 2,500 retained images. Of these, 1,750 images are used for the training set and 750 images for the testing set, and their locations are annotated accordingly.\\
\indent
\textbf{FIRESENSE}~\cite{42} comprises 49 videos. The processing methodology adopted in this study aligns with that of VisiFire, involving the retention of 2,500 images for analysis. Regression labels are assigned to those images.\\
\indent
\textbf{BoWFire}~\cite{43} consists of 466 images. However, the 240 images in the training set have low resolution. Consequently, they are removed, and only 226 images are retained. Of these, 158 images are used for the training set and 68 images for the testing set, with appropriate labeling.\\
\indent
\textbf{MS-FSDB}~\cite{48} pulished in 2024, consists of 12,586 images, with 8,983 images for training, 3,528 images for testing, and 11,327 images for the trainval subset. To ensure experimental fairness, 3,000 images from MS-FSDB are randomly selected as miniMS-FSDB.\\
\noindent
\textbf{Task.} The FSD task includes two subtasks: fire detection and smoke detection, each of which is performed on multiple public FSD datasets.\\
\noindent
\textbf{Metrics.} Uniform evaluation criteria are established and the 11-point interpolation method is utilized~\cite{37} to accurately measure the performance of the model. Note, \textbf{Fire} represents the Average Precision ($\%$) of fire, \textbf{Smoke} represents the Average Precision ($\%$) of smoke, and $\mathrm{mAP}$ represents the mean Average Precision ($\%$) of fire and smoke.\\
\noindent
\textbf{Implementation Details.} To ensure consistency and comparability, all model codes are standardized, and all datasets are uniformly preprocessed and converted to standardized formats. The experiments are performed on a computing platform equipped with an NVIDIA RTX 3090 Ti GPU to ensure efficient computational capabilities and accurate experimental results.\\
\indent
\subsection{Comparisons with State-of-the-Arts}
\textbf{Setting.} The baseline comprises four methods: SSD~\cite{18}, RetinaNet~\cite{45}, Faster RCNN~\cite{16}, and FCOS~\cite{36}, while the state-of-the-art (SOTA) approaches include Yolov5~\cite{26}, Yolov8~\cite{47}, and FCOS~\cite{36}. To ensure fairness, the input image sizes of different models are set within a similar range. The input image sizes for SSD are $300~\times~300$ for small (s) and $512~\times~512$ for large (l). In RetinaNet, the input image sizes are about $360~\times~360$ for small and about $608~\times~608$ for large. In Faster RCNN, the input image sizes are around $400~\times~666$ for small and around $800~\times~1,333$ for large. As for FCOS, the input image sizes are roughly $416~\times~608$ for small and roughly $832~\times~1,216$ for large. For YOLOv5 and YOLOv8, pixel values of $640~\times~640$ are used. In a-FSDM ,the input image sizes are roughly $416~\times~608$ for small and about $832~\times~1,216$ for large.\\
\noindent
\textbf{Main Results.} According to Table~\ref{tab1}, the comparative experimental results across nine FSD datasets indicate that the a-FSDM model consistently achieves the highest mean Average Precision (mAP) among all compared models.  Moreover, compared to the average mAP on each FSD dataset, our model demonstrates a significant improvement in precision. It is noteworthy that the baseline models SSD, RetinaNet, Faster RCNN, and FCOS achieve mAP scores of 89.4, 89.6, 96.6, and 96.3, respectively, on MS-FSDB with large-sized images. This suggests that, when trained and tested on the same dataset, anchor-based one-stage object detection algorithms generally perform worse in terms of accuracy compared to anchor-based two-stage and anchor-free detection algorithms. This performance gap is likely due to the fact that anchor-based one-stage detection algorithms generate a large number of anchors during computation, which inevitably introduces background noise. In contrast, anchor-based two-stage detection algorithms adopt different strategies to effectively mitigate background noise, thereby improving detection accuracy.\\
\indent
In addition, the baseline models SSD (mAP 77.8) and RetinaNet (mAP 84.9) on miniMS-FSDB with small-sized images do not outperform the SSD (mAP 84.1) and RetinaNet (mAP 90.7) models on the Fire-Smoke-Dataset. This is likely because, when the number of images in miniMS-FSDB is small and the targets are complex, the models may struggle to effectively extract target features, leading to reduced performance and even overfitting. In contrast, when the number of images in MS-FSDB is larger and the targets are complex, the models have access to a more diverse and representative set of samples, which enhances their generalization capability on MS-FSDB. It is important to point out that the BoWFire dataset contains only 226 images, and its lower precision may be due to overfitting caused by the limited number of images in this dataset.\\
\indent
Moreover, various FSD models are compared on miniMS-FSDB, and small input images are uniformly chosen for all FSD models to ensure consistency and comparability in experiments. By analyzing Table~\ref{tab3} in depth, the following important findings can be observed: a-FSDM outperforms other models in terms of mAP in Table~\ref{tab3}. Specifically, the proposed method’s AP for Fire reaches 98.3, and for Smoke, the AP also demonstrates strong performance at 99.3. Overall, the proposed method’s mAP is 98.8, which is notably high. Based on the analysis results, the proposed model demonstrates good adaptability and robustness.\\
\indent

\begin{sidewaystable*}
        \vspace{57em}
        \begin{minipage}{\textwidth}
        \caption{Baseline model comparison across different datasets. Fire, Smoke and $\mathrm{mAP}$ are given in the subsetion “Setting and Details”. “\textcolor{blue}{avg}” represents the average of mAP (mean Average Precision) values of all models across the FSD datasets. “s" represents the input image of small size, while “l" represents the input image of large size. F-RCNN means Faster RCNN.}
        \vspace{-1em}
        \label{tab1}
        \begin{tabularx}{\textwidth}{XXXXX | XXX | XXX | XXX |XXX |X}
            \toprule
            \diagbox{Dataset}{AP ($\%$)}{Model}& & \multicolumn{3}{c}{SSD\textsuperscript{\cite{18}}(s/l)} & \multicolumn{3}{c}{RetinaNet\textsuperscript{\cite{45}}(s/l)}&
            \multicolumn{3}{c}{F-RCNN\textsuperscript{\cite{16}}(s/l)}& \multicolumn{3}{c}{FCOS\textsuperscript{\cite{36}}(s/l)}& \multicolumn{3}{c}{a-FSDM(s/l)} & \textcolor{blue}{avg}\\
            
            \midrule
            && \rmfamily{Fire}& \rmfamily{Smoke}& $\mathrm{mAP}$& \rmfamily{Fire}& \rmfamily{Smoke}& $\mathrm{mAP}$& \rmfamily{Fire}& \rmfamily{Smoke}& $\mathrm{mAP}$& \rmfamily{Fire}& \rmfamily{Smoke}& $\mathrm{mAP}$&
            \rmfamily{Fire}& \rmfamily{Smoke}& $\mathrm{mAP}$&$\mathrm{mAP}$\\
            \multicolumn{2}{l}{Fire-Smoke-Dataset\textsuperscript{\cite{39}}(s)}&77.5&90.8&84.1&
            90.7&90.7&90.7&97.3&93.2&95.3&97.2&98.5&97.8&\textbf{97.5}&\textbf{99.2}&\textbf{98.3}& \textcolor{red}{93.2}\\
            \multicolumn{2}{l}{Furg-Fire-Dataset\textsuperscript{\cite{40}}(s)}&75.8&86.9&81.4&
            81.7&90.0&85.8&\textbf{95.8}&93.4&94.6&93.7&\textbf{98.1}&95.9&94.2&\textbf{98.1}&\textbf{96.1}& \textcolor{red}{90.8}\\
            \multicolumn{2}{l}{VisiFire\textsuperscript{\cite{41}}(s)}&78.2&89.5&83.9&
            84.2&90.7&87.4&92.8&88.8&90.8&88.8&96.7&92.8&\textbf{96.2}&\textbf{99.4}&\textbf{97.8}& \textcolor{red}{90.5}\\
            \multicolumn{2}{l}{FIRESENSE\textsuperscript{\cite{42}}(s)}&89.0&90.4&89.7&
            90.9&90.9&90.9&96.8&95.8&96.3&96.1&96.1&96.1&\textbf{98.3}&\textbf{98.1}&\textbf{98.2}& \textcolor{red}{94.2}\\
            \multicolumn{2}{l}{BoWFireDataset\textsuperscript{\cite{43}}(s)}&69.6&84.7&77.1&
            72.3&88.4&80.3&\textbf{86.3}&95.0&90.6&85.1&95.2&90.2&86.1&\textbf{97.9}&\textbf{92.0}& \textcolor{red}{86.0}\\
            \midrule
            \multicolumn{2}{l}{miniMS-FSDB(s)}&71.2&84.4&77.8&
            80.4&89.4&84.9&98.0&93.0&95.5&94.1&95.9&95.0&\textbf{98.3}&\textbf{99.3}&\textbf{98.8}& \textcolor{red}{90.4}\\
            \multicolumn{2}{l}{MS-FSDB(s)}&81.0&90.2&85.6&
            81.0&90.5&85.8&\textbf{98.2}&93.5&95.8&95.6&98.4&97.0&97.1&\textbf{98.9}&\textbf{98.0}& \textcolor{red}{92.4}\\
            \multicolumn{2}{l}{miniMS-FSDB(l)}&75.9&87.1&81.5&
            87.0&88.4&87.7&98.0&94.1&96.1&95.5&96.2&95.8&\textbf{98.1}&\textbf{97.6}&\textbf{97.9}& \textcolor{red}{91.8}\\
            \multicolumn{2}{l}{MS-FSDB(l)}&88.0&90.9&89.4&
            89.5&89.6&89.6&97.3&95.9&96.6&96.0&96.7&96.3&\textbf{98.4}&\textbf{98.6}&\textbf{98.5}& \textcolor{red}{94.1}\\
            \bottomrule
        \end{tabularx}
        \end{minipage}
        \begin{minipage}{\textwidth} 
        \caption{Comparison between generic detection heads and the Attention Transparency Detection Head (ATDH) across the MS-FSDB. Fire, Smoke and $\mathrm{mAP}$ are given in the subsetion “Setting and Details”. “s" represents the input image of small size, while “l" represents the input image of large size.}
        \vspace{-1em}
        \label{tab2}
        \begin{tabularx}{\textwidth}{XXXXX|XXX|XXX|XXX}
            \toprule
           \diagbox{Model}{AP}{Dataset}&&
            \multicolumn{3}{c}{miniMS-FSDB(s)}&
             \multicolumn{3}{c}{MS-FSDB(s)}&
             \multicolumn{3}{c}{miniMS-FSDB(l)}&
             \multicolumn{3}{c}{MS-FSDB(l)}\\ 
             \midrule
            &&\rmfamily{Fire}&\rmfamily{Smoke}&$\mathrm{mAP}$& \rmfamily{Fire}&\rmfamily{Smoke}&$\mathrm{mAP}$& \rmfamily{Fire}&\rmfamily{Smoke}&$\mathrm{mAP}$& \rmfamily{Fire}&\rmfamily{Smoke}&$\mathrm{mAP}$\\
            \multicolumn{2}{l}{SSD\textsuperscript{\cite{18}}}&71.2&84.4&77.8&81.0&90.2&85.6&75.9&87.1&81.5&88.0&\textbf{90.9}&89.4\\
            \multicolumn{2}{l}{~+ATDH} &\textbf{89.8}&\textbf{90.9}&\textbf{90.3}&\textbf{87.7}&\textbf{90.9}&\textbf{89.3}&\textbf{90.5}&\textbf{90.9}&\textbf{90.7}&\textbf{89.6}&90.8&\textbf{90.2}\\
            \midrule
            \multicolumn{2}{l}{RetinaNet\textsuperscript{\cite{45}}}&80.4&89.4&84.9&81.0&90.5&85.8&87.0&88.4&87.7&89.5&89.6&89.6\\
            \multicolumn{2}{l}{~+ATDH}   &\textbf{87.8}&\textbf{90.9}&\textbf{89.3}&\textbf{81.8}&\textbf{90.9}&\textbf{86.4}&\textbf{90.4}&\textbf{90.0}&\textbf{90.2}&\textbf{90.8}&\textbf{90.9}&\textbf{90.7}\\
            \midrule
            \multicolumn{2}{l}{Faster RCNN\textsuperscript{\cite{16}}}&98.0&93.0&95.5&98.2&93.5&95.8&98.0&94.1&96.1&97.3&95.9&96.6\\
            \multicolumn{2}{l}{~+ATDH}&\textbf{98.2}&\textbf{98.1}&\textbf{98.2}&\textbf{98.3}&\textbf{96.8}&\textbf{97.5}&\textbf{98.6}&\textbf{98.3}&\textbf{98.4}&\textbf{99.2}&\textbf{98.2}&\textbf{98.7}\\
            \midrule
            \multicolumn{2}{l}{FCOS\textsuperscript{\cite{36}}} &94.1&95.9&95.0&95.6&98.4&97.0 &95.5&96.2&95.8&96.0&96.7&96.3\\
        \multicolumn{2}{l}{~+ATDH} &\textbf{98.3}&\textbf{99.3}&\textbf{98.8}&\textbf{97.1}&\textbf{98.9}&\textbf{98.0}&\textbf{98.1}&\textbf{97.6}&\textbf{97.9}&\textbf{98.4}&\textbf{98.6}&\textbf{98.5}\\
        \bottomrule 
    \end{tabularx}
    \end{minipage}
    
    \begin{minipage}{\textwidth}
    \caption{The performance of different models. It shows in detail the performance of various seven detection models in terms of key metrics such as detection accuracy, $\mathrm{mAP}$, Performance (Gflops) and number of parameters (M) for each category, where Fire, Smoke and $\mathrm{mAP}$ are given in the subsetion “Setting and Details”. And, we use the input image of small size used the miniMS-FSDB. In models,~\cite{26} represents Yolov5,~\cite{47} stands for Yolov8,~\cite{18} denotes SSD,~\cite{45} signifies RetinaNet,~\cite{16} means Faster RCNN,~\cite{36} refers to FCOS, and ours represents the proposed a-FSDM model. Additionally, s, x, n, l, m represent different versions of the model.}
    \vspace{-1em}
    \label{tab3}
    \begin{tabularx}{\textwidth}{XXXXXXXXXXXXXXXXX}
    \toprule
    \multicolumn{2}{l}{Model }&5s\textsuperscript{\cite{26}}&5x\textsuperscript{\cite{26}}&5n\textsuperscript{\cite{26}}&5l\textsuperscript{\cite{26}}  &5m\textsuperscript{\cite{26}}&8s\textsuperscript{\cite{47}}&8x\textsuperscript{\cite{47}}&8n\textsuperscript{\cite{47}}&8l\textsuperscript{\cite{47}}&8m\textsuperscript{\cite{47}}&\cite{18}&\cite{45}&\cite{16}&\cite{36}&Ours\\
    \midrule
    \multicolumn{2}{l}{Fire ($\%$)}&66.3&67.5&72.4&71.5&69.6&87.9&74.4&83.6&72.5&84.3&71.2&80.4&98.0&94.1&\textbf{98.3}\\
    \multicolumn{2}{l}{Smoke ($\%$)}&81.3& 81.6&78.2&79.7&84.1&89.7&71.3&82.7&70.2&84.3&84.4&89.4&93.0&95.9&\textbf{99.3}\\
    \multicolumn{2}{l}{$\mathrm{mAP}$ ($\%$)}&73.8&74.5&75.3&75.6&76.8&88.8&72.8&83.1&71.3&84.4&77.8&84.9&95.5&95.0&\textbf{98.8}\\
    \multicolumn{2}{l}{Perf (Gflops)}&15.8&203.8&4.1&107.7&47.9&28.4&257.4&8.1&164.8&78.7&731.5&\textbf{1,715.3}&504.3&1,289.9&1,290.5\\
    \multicolumn{2}{l}{Params (M)}&7.0&86.2&\textbf{1.8}&46.1&20.9&11.1&68.1&2.9&41.6&24.6&23.9&36.4&41.4&32.1&33.3\\
    \bottomrule
    \end{tabularx}
    \end{minipage}
   
\end{sidewaystable*}

\subsection{Ablation Studies}
\begin{figure}[!t]
     \vspace{-1em}
      \centering
      \subfloat[]{\includegraphics[width=0.49\linewidth]{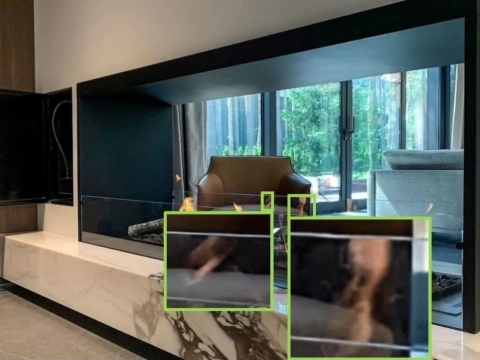}}
      \vspace{1pt}
      \subfloat[]{\includegraphics[width=0.49\linewidth]{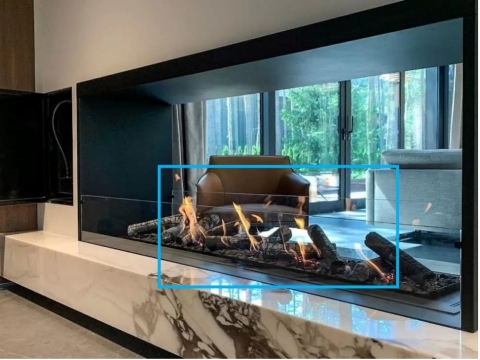}}
      \vspace{-1em}
      \caption{An image of containing transparent flame, (a) an enlarged patch of Transparent Background (flame) and Foreground (sofa). (b) Detection Result of the state-of-the-art generic detector~\cite{36}. The blue box in (b) represents ground truth, and the phenomenon of missed detection occurs in (b).}
      \vspace{-1em}
      \Description{}
      \label{fig2}
\end{figure}

\begin{figure}
    \subfloat[]{
    \includegraphics[width=0.24\linewidth]{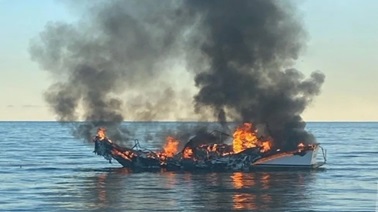}}
    \subfloat[]{
    \includegraphics[width=0.24\linewidth]{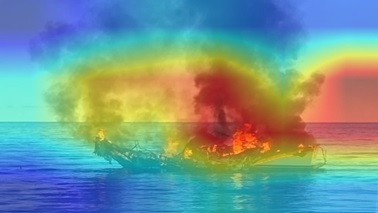}}
    \subfloat[]{
    \includegraphics[width=0.24\linewidth]{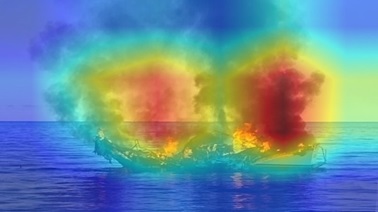}}
    \subfloat[]{
    \includegraphics[width=0.24\linewidth]{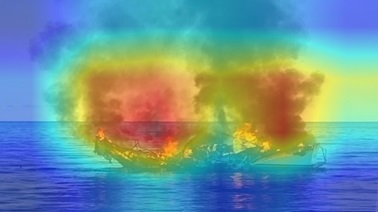}}
    \vspace{-1em}
    \caption{The evidence supporting the efficacy of the ATDH, (a) original image, (b) CAM visualization results of FCOS with SENet~\cite{44}, (c) CAM visualization results of FCOS with SKNet~\cite{34}. (d) CAM visualization results of FCOS with ATDH.}
    \vspace{-1em}
      \Description{}
      \label{fig3}    
\end{figure}

\begin{figure}
    \vspace{-1em}
    \subfloat[]{
    \includegraphics[width=0.24\linewidth]{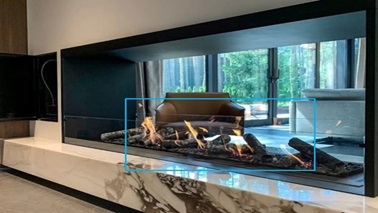}
    \includegraphics[width=0.24\linewidth]{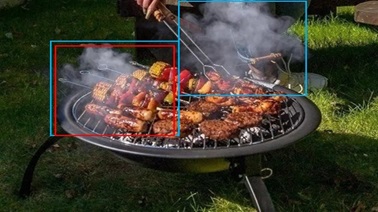}
    \includegraphics[width=0.24\linewidth]{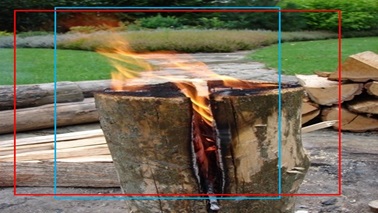}
    \includegraphics[width=0.24\linewidth]{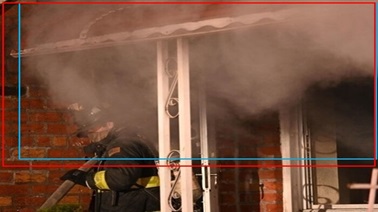}}
    \\
    \subfloat[]{
    \includegraphics[width=0.24\linewidth]{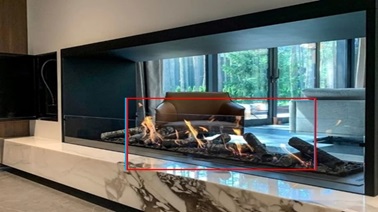}
    \includegraphics[width=0.24\linewidth]{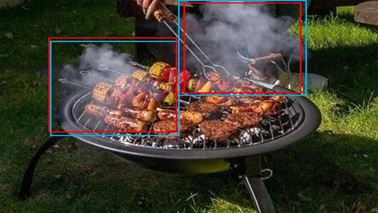}
    \includegraphics[width=0.24\linewidth]{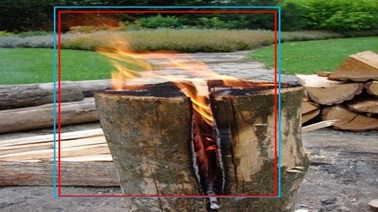}
    \includegraphics[width=0.24\linewidth]{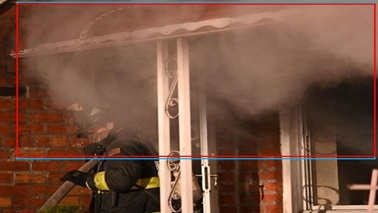}}
    \vspace{-1em}
    \caption{The Detection of Transparent Targets Images in FSD, (a) the false results of generic detection, (b) that the proposed method successfully detected the previous failure result. In the diagram, blue boxes represent ground truth and red boxes represent predicted results.}
    \Description{}
    \label{fig4}    
    \vspace{-1em}
\end{figure}

\textbf{Improves Performance with Transparency.} As discussed in the section “Introduction”, transparent flames or smoke frequently present in fires can result in inadequate detection or false alarms. Typically, the RGB values of the flame and the sofa are not similar. However, due to the transparency of the flame, its RGB values become similar to those of the sofa, both exhibiting a brown tone, specifically (55,32,19) and (60,14,124) respectively in Fig.~\ref{fig2} (a). Subsequently, when applying the state-of-the-art generic detector~\cite{36}, depicted in Fig.~\ref{fig2} (b), this transparent target suffers from missed detection.\\
\indent
The Attention Transparency Detection Head (ATDH) we designed proves to be highly effective. Initially, integrating the ATDH method into various baselines results in an improvement of at least 0.6 mAP (e.g., RetinaNet on MS-FSDB with small-sized images), as shown in Table~\ref{tab2}. Additionally, we conduct comparative experiments using SKNet~\cite{34} and SENet~\cite{44} on (mini) MS-FSDB with small-sized images, as presented in Table~\ref{tab4}. The experimental results indicate that ATDH significantly outperforms the general detection head with attention mechanisms, achieving a mAP increase of at least 1.0 on the MS-FSDB. Furthermore, in scenarios where the foreground exhibits partial transparency, as depicted in Fig.~\ref{fig3}, the CAM map~\cite{33, he2023mitigating} demonstrates that SENet~\cite{44} CAM assigns high values to non-smoke regions, and SKNet~\cite{34} CAM assigns values to non-fire regions. In contrast, the ATDH method effectively concentrates on the FSD target. Moreover, the visual results of detecting transparent fire and smoke, as shown in Fig.~\ref{fig4}, further validate the effectiveness of the ATDH method compared to a generic detection model. In the first and second images (from left to right), our method successfully identifies false detections made by the previous generic model, which arise due to the similarity between the transparent foreground and the background in the RGB image. In the third and fourth images, we observe lower regression errors, further highlighting the superior performance of the ATDH method.\\

\vspace{-1em}
\begin{table}[!htb]
    \caption{The attention mechanism algorithm added to the baseline (FCOS) on the MS-FSDB. Fire, Smoke and $\mathrm{mAP}$ are given in the subsetion “Setting and Details”. the input image of small size is used.}
    \vspace{-1em}
        \label{tab4}
        \renewcommand{\arraystretch}{0.35}
        \resizebox{0.45\textwidth}{!}{
        \begin{tabularx}{0.5\textwidth}{XXXXX|XXX}
            \toprule
            \diagbox{Model}{AP}{Dataset}&&
            \multicolumn{3}{c}{miniMS-FSDB}&\multicolumn{3}{c}{MS-FSDB}\\ 
             \midrule
             &&\rmfamily{Fire}&\rmfamily{Smoke}&$\mathrm{mAP}$&\rmfamily{Fire}&\rmfamily{Smoke}&$\mathrm{mAP}$\\
            \multicolumn{2}{l}{FCOS\textsuperscript{\cite{36}}} &94.1&95.9&95.0&95.6&98.4&97.0\\
            \multicolumn{2}{l}{~+SENet\textsuperscript{\cite{44}}} &95.9&98.1&97.0&94.7&98.0&96.3\\
            \multicolumn{2}{l}{~+SKNet\textsuperscript{\cite{34}}}&97.0&98.3&97.7&95.9&98.3&97.1\\
             \multicolumn{2}{l}{~+ATDH} &\textbf{98.3}&\textbf{99.3}&\textbf{98.8}&\textbf{97.1}&\textbf{98.9}&\textbf{98.0}\\
        \bottomrule 
        
    \end{tabularx}}
    \vspace{-1em}
\end{table}

\noindent
\textbf{Featuring Burning Intensity as Risk Assessment.} This is the first instance of burning intensity being employed as an indicator to quantify the severity of combustion. To further empirically validate the accuracy of burning intensity, we design an experiment, as outlined in Table~\ref{tab5}, and use the average burning intensity (avg BI) to measure each FSD dataset. The results of the experiment demonstrate that the average burning intensity (avg BI) of the FIRESENSE is markedly higher than that of the Fire-Smoke-Dataset. This discrepancy can be attributed to the fact that the FIRESENSE is of a larger scale, with the majority of fire scenes occurring in outdoor or forest settings.  In contrast, the Fire-Smoke-Dataset is of a smaller scale and predominantly encompasses indoor fire scenarios.  These findings suggest that our burning intensity metric is both effective and reasonable across a range of scenes.\\

\vspace{-1em}
\begin{table}[!htb]
    \caption{The reuslts of Burning Intensity ($\%$).~\cite{18} represents SSD,~\cite{45} indicates RetinaNet,~\cite{16} means Faster RCNN, and ~\cite{36} refers to FCOS, Ours represents the proposed a-FSDM model. “\textcolor{blue}{avg}” represents the average of burning intensity values of all models across the each FSD datasets.}
    \vspace{-1em}
    \label{tab5}
    \renewcommand{\arraystretch}{0.35}
    \resizebox{0.45\textwidth}{!}{
    \begin{tabularx}{0.5\textwidth}{XXXXXXX}
    \toprule
    \diagbox{Dataset}{BI($\%$)}{Model}&\cite{18}&\cite{45}&\cite{16}&\cite{36}&Ours&\textcolor{blue}{avg}\\
    \midrule
       \multicolumn{1}{l}{Fire-Smoke-Dataset\textsuperscript{\cite{39}}}&73.3&7.5&43.0&8.5&8.6&\textcolor{red}{28.2}\\
       \multicolumn{1}{l}{Furg-Fire-Dataset\textsuperscript{\cite{40}}}&64.7&14.0&49.9&8.7&10.3&\textcolor{red}{29.5}\\
       \multicolumn{1}{l}{VisiFire\textsuperscript{\cite{41}}}&72.1&14.1&65.5&10.2&9.6&\textcolor{red}{34.3}\\
       \multicolumn{1}{l}{FIRESENSE\textsuperscript{\cite{42}}}&81.4&50.3&68.2&7.0&11.2&\textcolor{red}{43.6}\\
       \multicolumn{1}{l}{BoWFireDataset\textsuperscript{\cite{43}}}&49.5&7.8&70.8&10.6&11.6&\textcolor{red}{30.1}\\
       \multicolumn{1}{l}{MS-FSDB}&53.6&22.0&72.0&8.9&9.4&\textcolor{red}{33.2}\\
    \bottomrule
    \end{tabularx}}
    \vspace{-1em}
\end{table}

\section{Conclusion}
The a-FSDM maintains the effective feature extraction and fusion capabilities of traditional detection algorithms while introducing a new ATDH to enhance performance and alleviate the problem of transparent fire or smoke object detection. Furthermore, burning intensity is innovatively proposed as an indicator of combustion severity. The a-FSDM and other generic object detection methods are validated using publicly available FSD datasets. Moreover, by comparing the proposed model against other generic object detection models, the viability and generalization capabilities of the proposed model are confirmed.\\
\indent

\begin{acks}
We would like to thank the anonymous reviewers for their insightful comments and suggestions. This work was supported by the National Natural Science Foundation of China (Grant No. U20B2066) and the Zhejiang Province "Pioneering Soldier" and "Leading Goose" R\&D Project (Grant No. 2023C01027).\\
\end{acks}
\bibliographystyle{ACM-Reference-Format}
\bibliography{references}


\end{document}